\documentclass[letterpaper, 10 pt, conference]{ieeeconf}  

\IEEEoverridecommandlockouts                              

\usepackage{graphics} 
\usepackage{graphicx}
\usepackage{lscape}
\usepackage{booktabs}
\usepackage{pgfplots}
\usepackage{pgfplotstable}

\title{\LARGE \bf
Emotional Musical Prosody for the Enhancement of Trust in Robotic Arm Communication
}

\author{Richard Savery, Lisa Zahray, Gil Weinberg$^{1}$ 
\thanks{$^{1}$Georgia Tech Center for Music Technology}
\thanks{\tt\small }}%

\begin{document}

\maketitle
\thispagestyle{empty}
\pagestyle{empty}

\begin{abstract}
As robotic arms become prevalent in industry it is crucial to improve levels of trust from human collaborators. Low levels of trust in human-robot interaction can reduce overall performance and prevent full robot utilization. We investigated the potential benefits of using emotional musical prosody to allow the robot to respond emotionally to the user's actions. We tested participants' responses to interacting with a virtual robot arm that acted as a decision agent, helping participants select the next number in a sequence. We compared results from three versions of the application in a between-group experiment, where the robot had different emotional reactions to the user's input depending on whether the user agreed with the robot and whether the user's choice was correct. In all versions, the robot reacted with emotional gestures. One version used prosody-based emotional audio phrases selected from our dataset of singer improvisations, the second version used audio consisting of a single pitch randomly assigned to each emotion, and the final version used no audio, only gestures. Our results showed no significant difference for the percentage of times users from each group agreed with the robot, and no difference between user's agreement with the robot after it made a mistake. However, participants also took a trust survey following the interaction, and we found that the reported trust ratings of the musical prosody group were significantly higher than both the single-pitch and no audio groups. 
\end{abstract}

\section{Introduction}
Robotic arms are showing a continual expansion in use, with expected growth continuing into the foreseeable future \cite{marketreport}. While the use of robotic arms continues to grow, they do not have a standard form of communication \cite{saunderson2019robots}, and current methods are costly to implement from both a technical and financial perspective. These systems generally focus on communicating intent, or describing what the arm will be about to perform, while less research has been done on the application of social and emotional communication. For collaborative processes, displaying emotion has repeatedly been shown to increase key collaboration metrics in robotics, such as likelihood of humans to follow social norms in robotic interactions\cite{jost2019examining}, better engagement with disability \cite{10.1145/3369457.3370915} and treating robots like an equal human collaborator \cite{desideri2019emotional}. It is even argued that no real collaboration can take place without social and emotional display \cite{fischer2019collaborative}.

We believe musical prosody, that uses non-linguistic audio phrases based on musical melodies, can be used for effective interaction with human collaborators without requiring a change in core functionality. The ability for sound to display information for robotic platforms beyond trivial indicators is often underutilized, despite the use of intentional sound to deliver information in almost every device we encounter day to day \cite{walker1996human}. It has also been shown that displaying emotion is key for creating believable agents that people enjoy collaborating with \cite{Mateas:1999:ORI:1805750.1805762}, and prosody is effective in displaying emotions for humans and robots \cite{crumpton2016survey}. While affective nonverbal behavior has been shown to affect HRI metrics like humans' emotional state, self-disclosure, and perceived animacy of the robot \cite{rosenthal2018effects}, gestures are often studied \cite{beck2010towards}, but non-linguistic forms of audio feedback are under-explored \cite{macdorman2006subjective}. Prosody has the potential to allow the robot to communicate in a manner relatable to that of humans, but still different enough from human speech to avoid the uncanny valley \cite{macdorman2006subjective}. Emotional musical prosody is therefore uniquely positioned to enable better robotic communication and collaboration, capturing the advantages of sonic interaction, emotion conveyance and avoiding uncanny valley. 

In this paper we describe our approach to musical prosody using a custom dataset of musical phrases for robotic arm interaction. We evaluate these interactions firstly to confirm that there is no impact through potential distraction in collaboration with a robotic arm. We then measure how musical prosody compares to single-pitch audio and no audio systems for trust, trust recovery, likeability and the perception of intelligence and safety.

\section{Background}
\subsection{Robotic Arm Forms of Communication}
Amongst research into methods for robotic arms to communicate and signal their intent, there is no standardized set of approaches \cite{cha2018survey}. In social robotics communication is often derived from human behaviour - such as gestures and gaze - these are not however readily available to robotic arms \cite{rosen2019communicating}. Additionally when these forms of communication are added to arms they require significant expense, such as extra visual displays like a face\cite{6839819}, or in the case of added gestures risk challenging and reducing the core functionality of the arm. In robotic research, forms of non-verbal communication can generally be split into four categories; kinesics, proxemics, haptics and chronemics, none of which are easily applied to an existing robotic system \cite{saunderson2019robots}. While varying movement to show intent has shown successful results \cite{bodden2016evaluating}, changes to path planning and movement dynamics is often not feasible. Another effective method for arms to display their intent is through vision of a robot's future trajectory, such as by a human worn head mounted display \cite{ruffaldi2016third}, however this requires a significant investment and potential distraction to the user.  Emotion has been more commonly used as an input to robotic arms, such as facial emotion recognition to control and change the behaviour of robotic arms \cite{saverysurvey}. Likewise, Galvanic Skin Response emotional evaluation on humans has been used to impact a robot's control pattern \cite{takahashi2001human}. Nevertheless, robotic arm displays of emotion in work and collaboration, or interaction beyond showing intent are widely overlooked in robotics literature.

\subsection{Communication for Trust and Trust Recovery}
For collaboration with robotic arms trust is required, without which they can be underutilized \cite{johndlee}. Trust is largely developed in the first phase of a relationship both between humans and robots \cite{kim2009repair}, meaning first impressions are crucial for trust. First impressions from audio and visual stimulus can also damage the ability to develop trust later on \cite{Schaefer2016}. In this work we focus on affective trust, which is developed through emotional bonds and personal relationship, not competence \cite{freedy2007measurement}. Affective trust makes relationships more resilient to mistakes by either party \cite{rousseau1998not}. The display of emotion is critical for affective trust and increases the willingness of collaboration \cite{gompei2018factors}.

Music and prosody has been shown as a powerful medium to convey emotions \cite{sloboda1999music}. In music and robotics emotion can be categorized in many ways, such as a discrete categorical manner (happiness, sadness, fear, etc.) \cite{devillers2005challenges}, and through continuous dimensions such as valence and arousal \cite{russell2009emotion}. Most recent efforts to generate and manipulate robotic emotions through prosody focused on linguistic robotic communication \cite{crumpton2016survey}.


\section{Methods}

\subsection{Research Questions and Hypotheses}



Our first research question focuses on understanding the role of musical prosody and trust for a robotic arm.

\begin{itemize}
    \item [RQ1] \textit{How  does emotional musical prosody  alter  trust and trust recovery from mistakes, compared to no audio and single-pitch audio?}
\end{itemize}

For this question our hypothesis is that the overall trust at the end of the interaction will be significantly higher for musical prosody over single-pitch and higher for single-pitch audio over no audio. 
Our next research question compares common HRI metrics between each system; the perceived intelligence, perceived safety and likeability of the robotic system.

\begin{itemize}
    \item [RQ2] \textit{How does emotional musical prosody alter perceived safety, perceived intelligence and likeability? }
\end{itemize}

Our second research question explores the relation between users' self-reported metrics, gathered through highly cited surveys and their actual responses collected through a performance based task. We are interested in comparing whether the system that is trusted more through self-reports is actually then utilized more in performance based tasks.

\begin{itemize}
    \item [RQ3] \textit{When a user indirectly self-reports higher levels of trust in a robot, does this in turn lead to higher utilization and trust in a robotic arm's suggestions?}
\end{itemize}

We hypothesize that users' self-reported trust ratings will correspond to their actual use and trust levels implied by choice to follow the decisions of the robotic system. We also hypothesize that by using musical prosody after mistakes, human collaborators will be more likely to trust the robotic arm's suggestions directly after a mistake. For the first two research questions, we believe that participants will develop an internal model of the robot as an interactive emotional collaborator for the prosody model. This will lead to higher levels of trust and improved perception of safety and intelligence.

\subsection{Experimental Design}
Our experiment requires participants to perform a pattern learning and prediction task collaboratively with a robot. This is followed by two commonly used surveys; Schaefer's survey for robotic trust \cite{schaefer2016measuring}, and the Godspeed measurement for Anthropomorphism, Animacy, Likeability, Perceived Intelligence, and the Perceived Safety of Robots\cite{bartneck2009measurement}.

The study process followed 5 steps for each participant:
\begin{enumerate}
    \item Consent form and introduction to online form
    \item Description of the pattern recognition task
    \item 20 Trial Pattern Recognition Tasks
    \item 80 Pattern Recognition Tasks, recorded for data 
    \item Godspeed and Schaefer Trust Survey (order randomized per participant)
\end{enumerate}

The pattern learning method was originally created by Dongen et al. to understand the reliance on decisions and develop a framework for testing different agents\cite{van2013framework}. Since then it has been re-purposed many times, including for comparing the dichotomy of human-human and human-automation trust \cite{de2012world}, as well as the use of audio by cognitive agents\cite{muralidharan2014effects}. 

\begin{figure} [t]
    \centering
    \includegraphics[width=5cm]{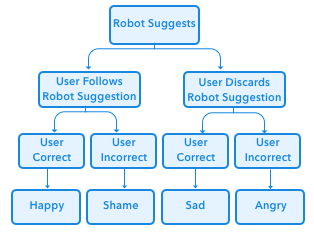}
    \caption{Robot Arm Emotional Response}
    \label{fig:sequence}
\end{figure}

In our version of the pattern recognition task, participants attempted to correctly predict the next number in a sequence. Participants were told beforehand that humans and the pattern recognition software being tested in the experiment tend to be about 70\% accurate on average, which has been shown to cause humans to alternate between relying on themselves and a decision agent. No further information was provided to the participants about the sequence's structure. The sequence was made up of a repeated sub-sequence that was 5 numbers long, containing only 1, 2, or 3 (such as 3, 1, 1, 2, 3). To prevent the ability for participants to quickly identify the pattern, 10\% of the numbers in the sequence were randomly altered. Participants first completed a training exercise to learn the interface, in which a sub-sequence was repeated 4 times (20 total numbers). Then participants were informed a new sequence had been generated for the final task. This was generated in the same way, using a new sub-sequence with 16 repetitions (80 total numbers). Before the user chose which number they believed came next in the sequence, the robot would suggest an answer, with the robot being correct 70\% of the time (see Figure \ref{fig:sequence}). This process mirrors the process from the original paper \cite{van2013framework}. 

The previous timestep's correct answer was displayed for the user at decision time to help them better keep track of the pattern throughout the time the robot takes to perform its movements. We required participants to submit their answer after the robot finished pointing to its prediction, which took between 2.5 and 4.5 seconds. This also forced participants to spend time considering their decision given the robot's recommendation. The robot would respond to the user's choice depending on the outcome and the version of the experiment, described in the following section.

\subsection{Experimental Groups and Robot Reactions}
\label{subsec:groups}
Our study was designed as a between-group experiment, where participants were randomly allocated to one of three groups. These groups were a prosody audio group (prosody), a single-pitch audio group (notes), and a control with no audio (gesture). The robot always responded to a user's action with the emotion determined by the process shown in Figure \ref{fig:sequence}. In all three versions of the experiment, the robot responded with the emotional gestures described in Section \ref{sec:gestures}. 

In the prosody group, the robot additionally responded with prosody-based audio sample, randomly selected each time from the five phrases matching the response emotion. These phrases which were obtained using the process described in Section \ref{sec:dataset}. In the notes group, the robot additionally responded instead with an audio file playing one note. Each emotion was randomly assigned one pitch from the midi pitches 62, 65, 69, and 72. This assignment remained consistent throughout the experiment to maintain a relation between the sounds and the outcome. For each pitch, five different audio files were available to be selected, each with a different instrument timbre and length (varying from 2-5 seconds), to provide variety similar to that of the five different prosody phrases available for each emotion. Finally, in the gesture group, the gesture was performed in silence.




\subsection{Participants}
We recruited 46 participants through the online survey platform Prolific\footnote{https://www.prolific.co/}. The participants ages ranged from 19 to 49, while the mean age was 25, with a standard deviation of 7. Participants were randomly sorted into one of the three categories, audio with emotional musical prosody (15 participants), single-pitch audio (16 participants), and no audio (15 participants). Each experiment took approximately 30 minutes to complete. Participants were paid \$4.75USD.

\subsection{Dataset} 
\label{sec:dataset}
\begin{figure} [t]
    \centering
    \includegraphics[width=7cm]{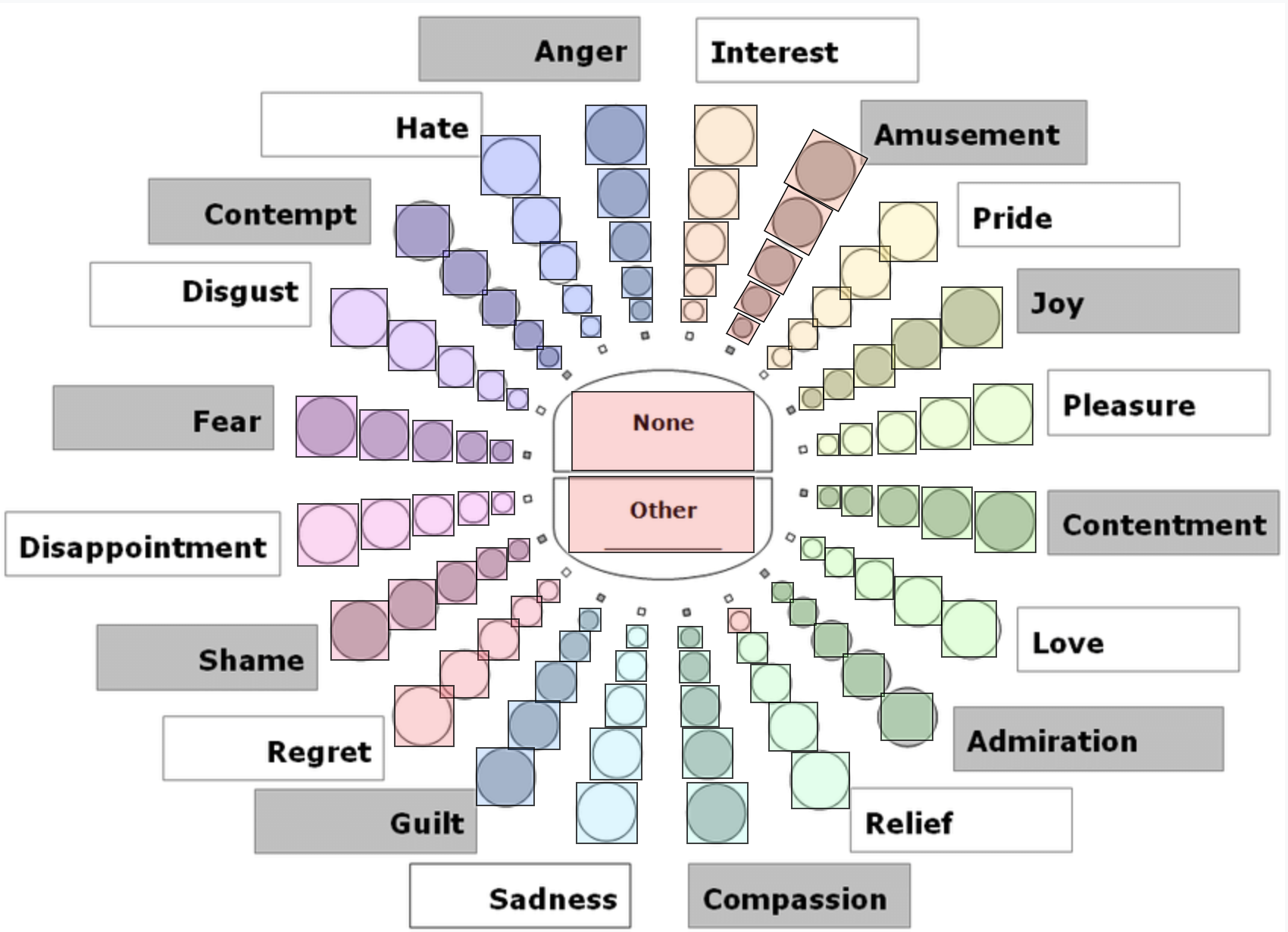}
    \caption{Geneva Emotion Wheel}
    \label{fig:geneva}
\end{figure}

In past work we created a deep learning generation system for musical prosody \cite{savery_finding_2019,savery2019establishing}. For this paper and experiment, we chose to use our recently created dataset of a human singing emotional phrases, to avoid any potential noise added by a generative system. The recorded dataset contains 4.22 hours of musical material recorded by Mary Carter\footnote{https://maryesthercarter.com/}, divided into 1-15 second phrases each corresponding to one of the 20 different emotions in the Geneva Emotion Wheel \cite{sacharin2012geneva} shown in Figure \ref{fig:geneva}.

As part of our evaluation of the dataset, we manually selected 5 phrases for each emotion that we felt best represented the emotion, and had participants select an emotion and intensity when listening to each provided phrase. Participants recruited from Prolific and MTurk were used. For quality assurance, participants were randomly given test questions throughout the experiment telling them to select a certain answer. Each participant was given 6.5 on average, and responses which had more than one incorrect attention question were ignored, leaving a total of 45 participants for data analysis.  In order to prevent the survey from being overly long, questions were randomly allocated, with 12 participants on average evaluating each individual phrase. Answers of None or Other were ignored in the analysis, resulting in an average of 11.3 valid evaluations for each phrase. 

Our analysis of the phrases used the metrics defined by Coyne et al\cite{coyne2020using}. We calculated the rated emotion's mean and variance in units of emotion (converted from degrees on the wheel), weighted by user-rated intensity. 

For the experiment, we used phrases for the four emotions joy, shame, sadness, and anger. These emotions were chosen as both being the emotions that best matched the outcomes in \ref{fig:sequence} and had gesture descriptions specified in \cite{walbott98}. 5 phrases for each emotion were chosen to add variety to the robot's response to not tire the user with the same sounds, while still allowing for only high-quality phrases to be included. In selecting the phrases for each of the four emotions, phrases from the closest two other emotions on the wheel within the same quadrant were also considered for selection. The sets were therefore \{joy, pride, pleasure\}, \{shame, disappointment, regret\},  \{sadness, guilt, regret\}, and \{anger, hate, contempt\}. We selected 5 of the 15 potential phrases for each by limiting length to be between 4 and 10 seconds, restricting the variance to be less than 2, requiring the weighted mean emotion rating to fall within the correct quadrant of the wheel, and finally selecting the phrases with the smallest difference between the actual emotion and mean rated emotion.


\subsection{Interaction} 
\begin{figure} [h]
    \centering
    \includegraphics[width=7cm]{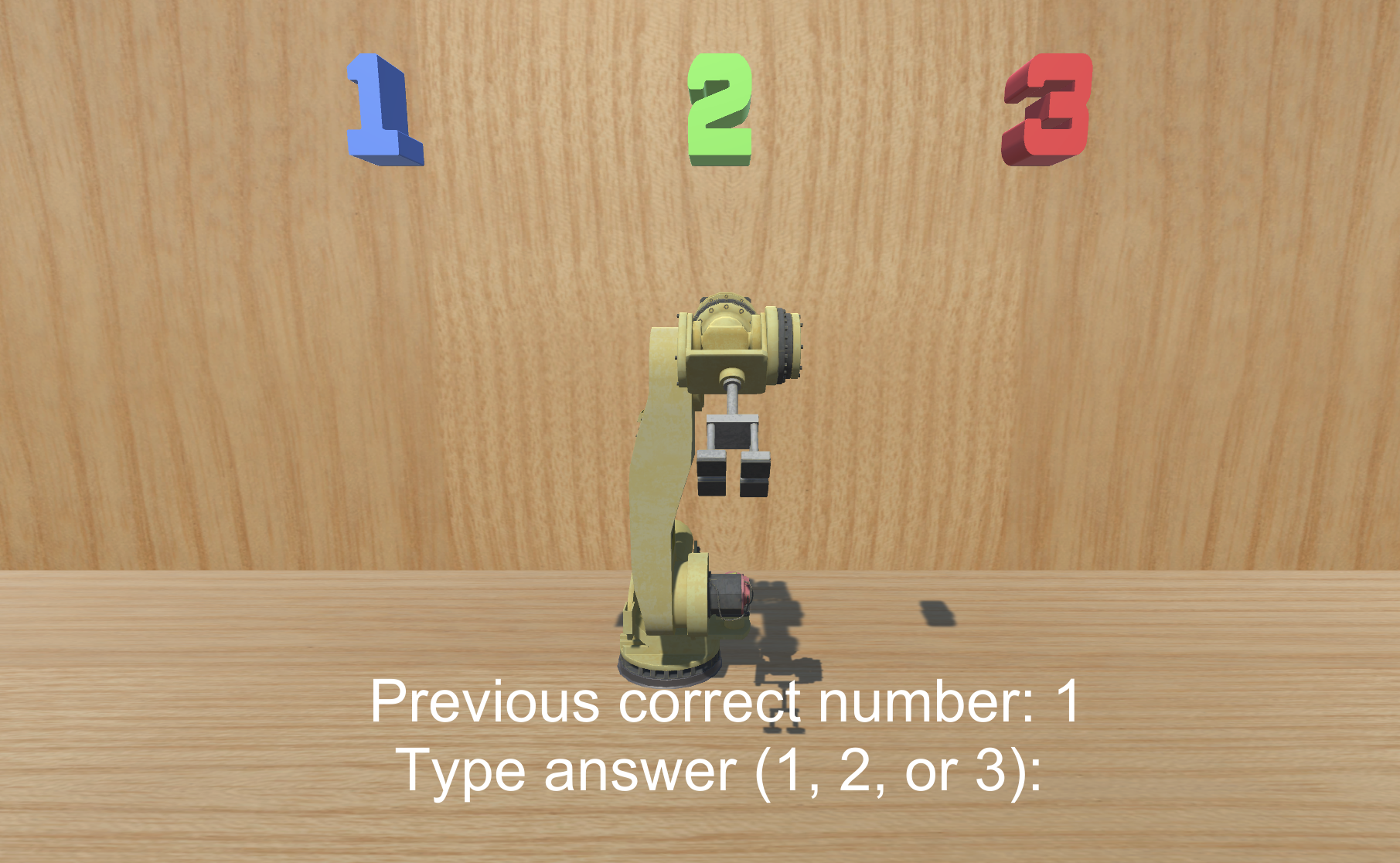}
    \caption{Example image from the robot interaction application}
    \label{fig:interface}
\end{figure}

Participants interacted with a virtual 3-D model of the robot in an application designed in Unity. Each time a participant was asked to answer a question, the robot acted as a decision agent, pointing to an answer that may be correct or incorrect. The user would then type their answer using their computer keyboard. There were three versions of the interaction application, varying the way the robot reacted to the user's answer, that are described in Section \ref{subsec:groups}. An example image of the interface is shown in Figure \ref{fig:interface}.

\subsection{Gestures}
\label{sec:gestures}
\begin{figure} [h]
    \centering
    \includegraphics[width=7cm]{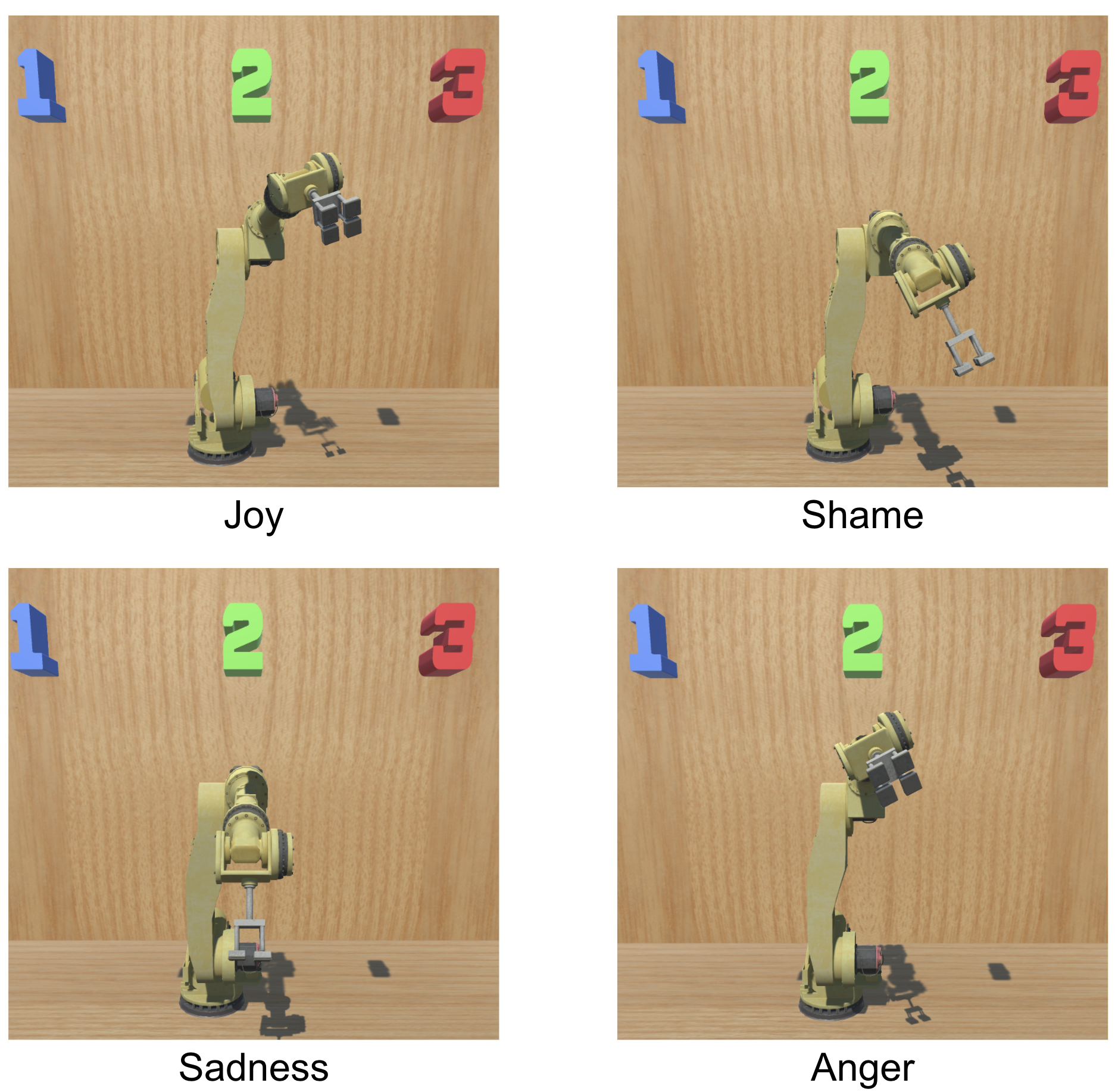}
    \caption{Example poses passed through during emotional gestures}
    \label{fig:gesturePoses}
\end{figure}

We created a gesture for each of the emotions joy, shame, sadness, and anger, as one way the robot reacted to user input. We designed the gestures by utilizing the table of emotion-specific nonverbal behaviors provided in \cite{walbott98}, which is based on work by Darwin, as well as their post-hoc overview of discriminative body movements and poses. These ideas have been used before in designing emotional robot gestures \cite{bretan2015emotionally}. Our joy gesture has the robot lift its arm up high, making three quick upwards movements alternating which side it faces. The shame gesture has the robot slowly bend down and away from the camera to one side. The sadness gesture has the robot slowly bend down while still centered with respect to the camera. The anger gesture has the robot first lean downwards and make two fast lateral movements, and then lean upwards to make two more fast lateral movements. Examples of poses encountered during each gesture are shown in Figure \ref{fig:gesturePoses}. 

\section{Results}

\begin{figure}
    \centering
    \includegraphics[width=6.5cm]{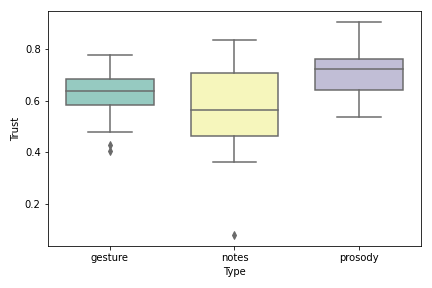}
    \caption{Box plot of trust scores}
    \label{fig:trust} 
\end{figure}

\begin{figure}
    \centering
    \includegraphics[width=6.5cm]{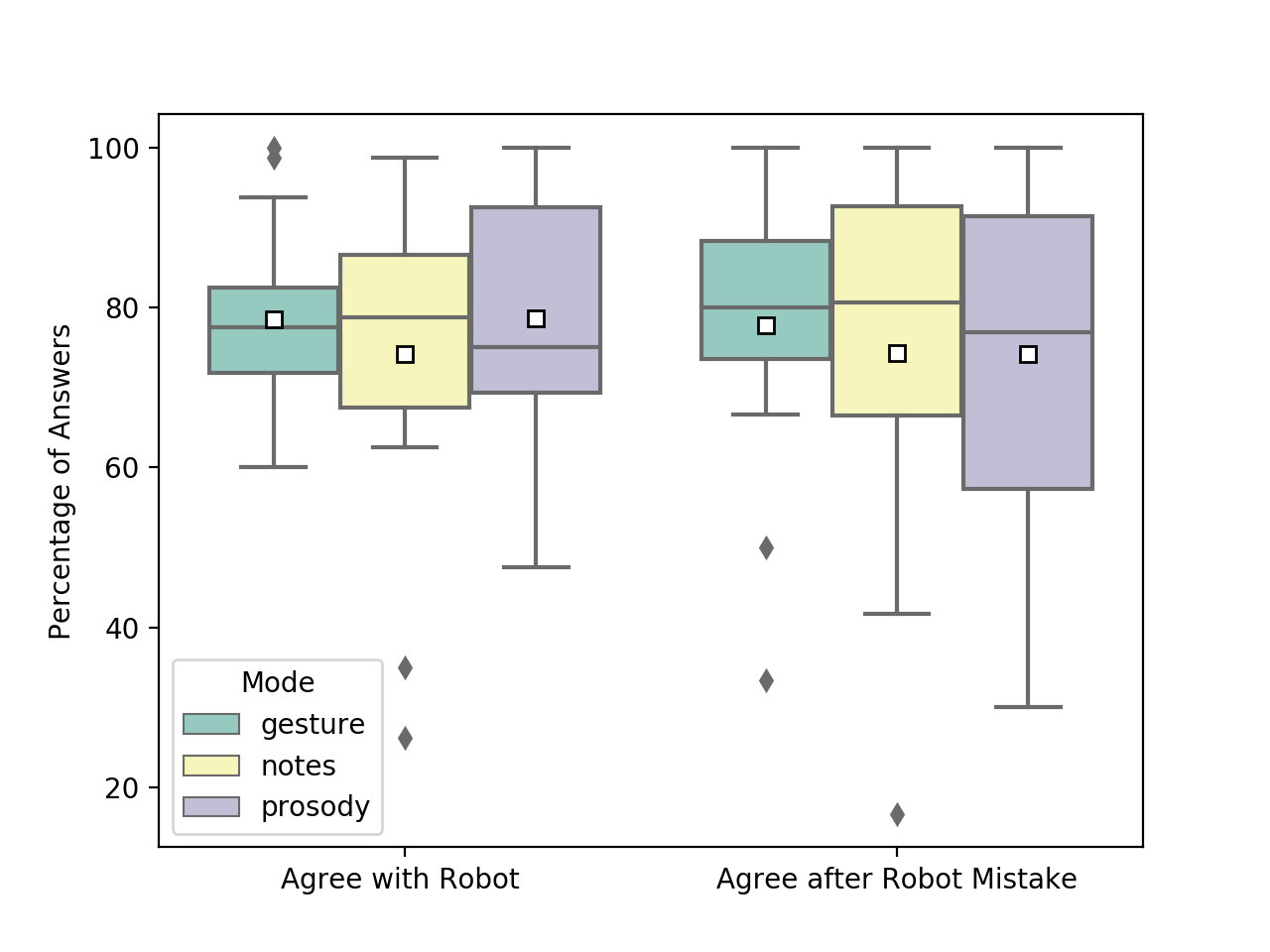}
    \caption{Box plot showing percentage of answers agreeing with the robot overall and after the robot made a mistake (means indicated by white squares)}
    \label{fig:agreeBox}
\end{figure}

\subsection{RQ1: Trust Recovery}
We first calculated Cronbach's alpha for each metric in the trust survey, which gave a high reliability of 0.92. We then calculated the overall trust score by inverting categories when appropriate and then generating the mean for each individual. The mean trust of each group was prosody 0.71, notes 0.57 and gesture 0.62 (see Figure \ref{fig:trust}). After running a one-way ANOVA the p-value was significant, \textit{p}=0.041. Pair-wise t-tests between groups' trust rating gave the results: notes-gestures \textit{p}= 0.46, notes-prosody \textit{p}=0.025, and gesture-prosody \textit{p}=0.025. This supports our hypothesis that trust would be higher from the arm using prosody. 

We also evaluated trust based on participants' actual use of the system. The percentage of answers for which users agreed with the robot for each group are plotted in Figure \ref{fig:agreeBox}. We performed a one-way ANOVA test to test whether there was a significant difference in this metric between groups, \textit{p}=0.68, which was not significant. 

To compare trust recovery after mistakes between groups, we analyzed the percentage of times each user agreed with the robot immediately after an instance of following the robot's incorrect suggestion. The results are plotted in Figure \ref{fig:agreeBox}. The one-way ANOVA test yielded \textit{p}=0.87, which was not significant.

\subsection{RQ2: Safety, Intelligence and Likeability}
\begin{figure}
    \centering
    \includegraphics[width=7cm]{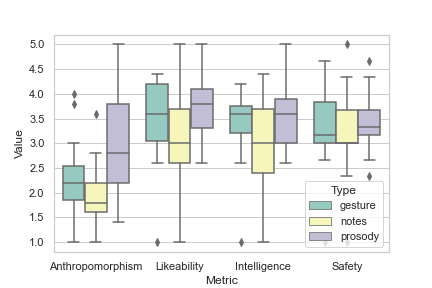}
    \caption{Box plot of HRI metrics}
    \label{fig:metrics}
\end{figure}
Cronbach's alpha for Anthropomorphism (0.85), Intelligence (0.89) and Likeability (0.92) all showed high reliability values above 0.85. Safety's coefficient was slightly lower at 0.75. Across each category results showed a higher median for each metric for the system using emotional prosody (see Figure \ref{fig:metrics}), while gestures consistently outperformed notes. We performed a one-way ANOVA on each category, and only Anthropomorphism was significant, \textit{p}=0.006.

\subsection{RQ3: Trust Survey and Participant Choices}
We calculated the Pearson correlation coefficient between the final trust scores, and the percentage of answers users agreed with the robot. The result was \textit{r}=0.12, which indicates a weak correlation between the two metrics. 

\subsection{User Comments} 
The comments provided by participants indicate that it was possible, in all groups, to perceive the emotions the robot was trying to convey. In the prosody group, one user said, `The arm seems quite emotional! When it's right it is quite happy, but when it is wrong it gets particularly sad.' In the notes group, a user said `When we got the right answer the robot seemed cheerful, as opposed to when we selected the wrong answer (based on the robot's recommendation) it seemed as if he was sorry for giving wrong suggestions. If I chose an option different than the robot's suggestion and its answer was correct, it seemed as if he gave the look of I told you the right answer!' And in the gesture group, one comment was `the emotions were very easily perceivable.' Two participants in the notes group had negative comments on the audio response, describing it as `horrible' and `annoying', while one participant in the prosody group said the `humming was annoying.' Several participants mentioned that the robot moved too slowly. Some comments mentioned having a hard time detecting any pattern in the sequence, while in others users discussed their strategies. 
\section{Discussion and Conclusion}
This study was performed using virtual interactions with a robot, and 46 participants. It would be useful to investigate this further with a larger sample size, and to have participants interact with a physical robot for comparison. Additionally, more variations of robot responses could be compared and analyzed beyond the three that we investigated. For example, prosodic audio of a human voice could be compared with that of musical instruments. 
Our results support that when the robot responded with musical prosody (alongside the gestures present for all groups), users reported higher trust metrics than when the robot responded with single-pitched notes or no audio. This supports that musical prosody has a positive effect on humans' trust of a robot.  
Comparing the Godspeed metrics, it was unsurprising to find that the addition of human vocalizations increased the Anthropomorphism of the system. We had expected likeability to be higher, and while it was not a significant result, it would still be worth investigating further with more subjects. The most surprising result was that the notes audio fell well below the median of gestures-only in every category. We believe this shows that while prosody can have positive outcomes, audio when implemented ineffectively has the capability to drastically reduce HRI metrics. The reason for this is likely due to the fact that the notes audio was not related to the emotion being displayed by the gesture beyond remaining consistent throughout the experiment. However, it would be interesting to further explore more types of audio responses. 

Users' ratings of trust in the survey did not strongly correlate with their actual behavior during the task, in terms of how often they agreed with the robot's suggestions. This is consistent with the fact that while users reported significantly higher trust for audio with musical prosody, no significant differences were found in their actual choices during the interactions. A similar conflict between these types of metrics was found in the original decision framework paper \cite{van2013framework}, where higher reported trust in the decision aid did not always result in higher percent agreement with the aid. Some potential explanations include cognitive biases and reliance heuristic.

\section{Acknowledgements}
This material is based upon work supported by the National Science Foundation under Grant No. 1925178 

\bibliographystyle{IEEEtran}
\bibliography{name}

\begin{thebibliography}{10}
\providecommand{\url}[1]{#1}
\csname url@samestyle\endcsname
\providecommand{\newblock}{\relax}
\providecommand{\bibinfo}[2]{#2}
\providecommand{\BIBentrySTDinterwordspacing}{\spaceskip=0pt\relax}
\providecommand{\BIBentryALTinterwordstretchfactor}{4}
\providecommand{\BIBentryALTinterwordspacing}{\spaceskip=\fontdimen2\font plus
\BIBentryALTinterwordstretchfactor\fontdimen3\font minus
  \fontdimen4\font\relax}
\providecommand{\BIBforeignlanguage}[2]{{%
\expandafter\ifx\csname l@#1\endcsname\relax
\typeout{** WARNING: IEEEtran.bst: No hyphenation pattern has been}%
\typeout{** loaded for the language `#1'. Using the pattern for}%
\typeout{** the default language instead.}%
\else
\language=\csname l@#1\endcsname
\fi
#2}}
\providecommand{\BIBdecl}{\relax}
\BIBdecl

\bibitem{marketreport}
{Grand View Research Choice}, ``Collaborative robots market size, share and
  trends analysis report by payload capacity, by application (assembly,
  handling, packaging, quality testing), by vertical, by region, and segment
  forecasts, 2019 - 2025,'' Grand View Research Choice, Tech. Rep.

\bibitem{saunderson2019robots}
S.~Saunderson and G.~Nejat, ``How robots influence humans: A survey of
  nonverbal communication in social human--robot interaction,''
  \emph{International Journal of Social Robotics}, vol.~11, no.~4, pp.
  575--608, 2019.

\bibitem{jost2019examining}
J.~Jost, T.~Kirks, S.~Chapman, and G.~Rinkenauer, ``Examining the effects of
  height, velocity and emotional representation of a social transport robot and
  human factors in human-robot collaboration,'' in \emph{IFIP Conference on
  Human-Computer Interaction}.\hskip 1em plus 0.5em minus 0.4em\relax Springer,
  2019, pp. 517--526.

\bibitem{10.1145/3369457.3370915}
\BIBentryALTinterwordspacing
S.~S. Balasuriya, L.~Sitbon, M.~Brereton, and S.~Koplick, ``How can social
  robots spark collaboration and engagement among people with intellectual
  disability?'' in \emph{Proceedings of the 31st Australian Conference on
  Human-Computer-Interaction}, ser. OZCHI’19.\hskip 1em plus 0.5em minus
  0.4em\relax New York, NY, USA: Association for Computing Machinery, 2019, p.
  209–220. [Online]. Available: \url{https://doi.org/10.1145/3369457.3370915}
\BIBentrySTDinterwordspacing

\bibitem{desideri2019emotional}
L.~Desideri, C.~Ottaviani, M.~Malavasi, R.~di~Marzio, and P.~Bonifacci,
  ``Emotional processes in human-robot interaction during brief cognitive
  testing,'' \emph{Computers in Human Behavior}, vol.~90, pp. 331--342, 2019.

\bibitem{fischer2019collaborative}
K.~Fischer, ``Why collaborative robots must be social (and even emotional)
  actors,'' \emph{Techn{\'e}: Research in Philosophy and Technology}, vol.~23,
  no.~3, pp. 270--289, 2019.

\bibitem{walker1996human}
B.~N. Walker and G.~Kramer, ``Human factors and the acoustic ecology:
  Considerations for multimedia audio design,'' in \emph{Audio Engineering
  Society Convention 101}.\hskip 1em plus 0.5em minus 0.4em\relax Audio
  Engineering Society, 1996.

\bibitem{Mateas:1999:ORI:1805750.1805762}
\BIBentryALTinterwordspacing
M.~Mateas, ``Artificial intelligence today,'' M.~J. Wooldridge and M.~Veloso,
  Eds.\hskip 1em plus 0.5em minus 0.4em\relax Berlin, Heidelberg:
  Springer-Verlag, 1999, ch. An Oz-centric Review of Interactive Drama and
  Believable Agents, pp. 297--328. [Online]. Available:
  \url{http://dl.acm.org/citation.cfm?id=1805750.1805762}
\BIBentrySTDinterwordspacing

\bibitem{crumpton2016survey}
J.~Crumpton and C.~L. Bethel, ``A survey of using vocal prosody to convey
  emotion in robot speech,'' \emph{International Journal of Social Robotics},
  vol.~8, no.~2, pp. 271--285, 2016.

\bibitem{rosenthal2018effects}
A.~M. Rosenthal-von~der P{\"u}tten, N.~C. Kr{\"a}mer, and J.~Herrmann, ``The
  effects of humanlike and robot-specific affective nonverbal behavior on
  perception, emotion, and behavior,'' \emph{International Journal of Social
  Robotics}, vol.~10, no.~5, pp. 569--582, 2018.

\bibitem{beck2010towards}
A.~Beck, L.~Ca{\~n}amero, and K.~A. Bard, ``Towards an affect space for robots
  to display emotional body language,'' in \emph{Ro-man 2010}.\hskip 1em plus
  0.5em minus 0.4em\relax IEEE, 2010, pp. 464--469.

\bibitem{macdorman2006subjective}
K.~F. MacDorman, ``Subjective ratings of robot video clips for human likeness,
  familiarity, and eeriness: An exploration of the uncanny valley,'' in
  \emph{ICCS/CogSci-2006 long symposium: Toward social mechanisms of android
  science}, 2006, pp. 26--29.

\bibitem{cha2018survey}
E.~Cha, Y.~Kim, T.~Fong, M.~J. Mataric \emph{et~al.}, ``A survey of nonverbal
  signaling methods for non-humanoid robots,'' \emph{Foundations and
  Trends{\textregistered} in Robotics}, vol.~6, no.~4, pp. 211--323, 2018.

\bibitem{rosen2019communicating}
E.~Rosen, D.~Whitney, E.~Phillips, G.~Chien, J.~Tompkin, G.~Konidaris, and
  S.~Tellex, ``Communicating and controlling robot arm motion intent through
  mixed-reality head-mounted displays,'' \emph{The International Journal of
  Robotics Research}, vol.~38, no. 12-13, pp. 1513--1526, 2019.

\bibitem{6839819}
W.~{Jitviriya} and E.~{Hayashi}, ``Design of emotion generation model and
  action selection for robots using a self organizing map,'' in \emph{2014 11th
  International Conference on Electrical Engineering/Electronics, Computer,
  Telecommunications and Information Technology (ECTI-CON)}, 2014, pp. 1--6.

\bibitem{bodden2016evaluating}
C.~Bodden, D.~Rakita, B.~Mutlu, and M.~Gleicher, ``Evaluating intent-expressive
  robot arm motion,'' in \emph{2016 25th IEEE International Symposium on Robot
  and Human Interactive Communication (RO-MAN)}.\hskip 1em plus 0.5em minus
  0.4em\relax IEEE, 2016, pp. 658--663.

\bibitem{ruffaldi2016third}
E.~Ruffaldi, F.~Brizzi, F.~Tecchia, and S.~Bacinelli, ``Third point of view
  augmented reality for robot intentions visualization,'' in
  \emph{International Conference on Augmented Reality, Virtual Reality and
  Computer Graphics}.\hskip 1em plus 0.5em minus 0.4em\relax Springer, 2016,
  pp. 471--478.

\bibitem{saverysurvey}
R.~Savery and G.~Weinberg, ``A survey of robotics and emotion: Classifications
  and models of emotional interaction,'' in \emph{Proceedings of the 29th
  International Conference on Robot and Human Interactive Communication}, 2020.

\bibitem{takahashi2001human}
Y.~Takahashi, N.~Hasegawa, K.~Takahashi, and T.~Hatakeyama, ``Human interface
  using pc display with head pointing device for eating assist robot and
  emotional evaluation by gsr sensor,'' in \emph{Proceedings 2001 ICRA. IEEE
  International Conference on Robotics and Automation (Cat. No. 01CH37164)},
  vol.~4.\hskip 1em plus 0.5em minus 0.4em\relax IEEE, 2001, pp. 3674--3679.

\bibitem{johndlee}
J.~D. Lee and K.~A. See, ``Trust in automation: Designing for appropriate
  reliance,'' \emph{Human Factors}, vol.~46, no.~1, pp. 50--80, 2004.

\bibitem{kim2009repair}
P.~H. Kim, K.~T. Dirks, and C.~D. Cooper, ``The repair of trust: A dynamic
  bilateral perspective and multilevel conceptualization,'' \emph{Academy of
  Management Review}, vol.~34, no.~3, pp. 401--422, 2009.

\bibitem{Schaefer2016}
K.~E. Schaefer, \emph{Measuring Trust in Human Robot Interactions: Development
  of the ``Trust Perception Scale-HRI''}.\hskip 1em plus 0.5em minus
  0.4em\relax Boston, MA: Springer US, 2016, pp. 191--218.

\bibitem{freedy2007measurement}
A.~Freedy, E.~DeVisser, G.~Weltman, and N.~Coeyman, ``Measurement of trust in
  human-robot collaboration,'' in \emph{Collaborative Technologies and Systems,
  2007. CTS 2007. International Symposium on}.\hskip 1em plus 0.5em minus
  0.4em\relax IEEE, 2007, pp. 106--114.

\bibitem{rousseau1998not}
D.~M. Rousseau, S.~B. Sitkin, R.~S. Burt, and C.~Camerer, ``Not so different
  after all: A cross-discipline view of trust,'' \emph{Academy of management
  review}, vol.~23, no.~3, pp. 393--404, 1998.

\bibitem{gompei2018factors}
T.~Gompei and H.~Umemuro, ``Factors and development of cognitive and affective
  trust on social robots,'' in \emph{International Conference on Social
  Robotics}.\hskip 1em plus 0.5em minus 0.4em\relax Springer, 2018, pp. 45--54.

\bibitem{sloboda1999music}
J.~Sloboda, ``Music: Where cognition and emotion meet,'' in \emph{Conference
  Proceedings: Opening the Umbrella; an Encompassing View of Music Education;
  Australian Society for Music Education, XII National Conference, University
  of Sydney, NSW, Australia, 09-13 July 1999}.\hskip 1em plus 0.5em minus
  0.4em\relax Australian Society for Music Education, 1999, p. 175.

\bibitem{devillers2005challenges}
L.~Devillers, L.~Vidrascu, and L.~Lamel, ``Challenges in real-life emotion
  annotation and machine learning based detection,'' \emph{Neural Networks},
  vol.~18, no.~4, pp. 407--422, 2005.

\bibitem{russell2009emotion}
J.~A. Russell, ``Emotion, core affect, and psychological construction,''
  \emph{Cognition and Emotion}, vol.~23, no.~7, pp. 1259--1283, 2009.

\bibitem{schaefer2016measuring}
K.~E. Schaefer, ``Measuring trust in human robot interactions: Development of
  the “trust perception scale-hri”,'' in \emph{Robust Intelligence and
  Trust in Autonomous Systems}.\hskip 1em plus 0.5em minus 0.4em\relax
  Springer, 2016, pp. 191--218.

\bibitem{bartneck2009measurement}
C.~Bartneck, D.~Kuli{\'c}, E.~Croft, and S.~Zoghbi, ``Measurement instruments
  for the anthropomorphism, animacy, likeability, perceived intelligence, and
  perceived safety of robots,'' \emph{International journal of social
  robotics}, vol.~1, no.~1, pp. 71--81, 2009.

\bibitem{van2013framework}
K.~Van~Dongen and P.-P. Van~Maanen, ``A framework for explaining reliance on
  decision aids,'' \emph{International Journal of Human-Computer Studies},
  vol.~71, no.~4, pp. 410--424, 2013.

\bibitem{de2012world}
E.~J. de~Visser, F.~Krueger, P.~McKnight, S.~Scheid, M.~Smith, S.~Chalk, and
  R.~Parasuraman, ``The world is not enough: Trust in cognitive agents,'' in
  \emph{Proceedings of the Human Factors and Ergonomics Society Annual
  Meeting}, vol.~56, no.~1.\hskip 1em plus 0.5em minus 0.4em\relax Sage
  Publications Sage CA: Los Angeles, CA, 2012, pp. 263--267.

\bibitem{muralidharan2014effects}
L.~Muralidharan, E.~J. de~Visser, and R.~Parasuraman, ``The effects of pitch
  contour and flanging on trust in speaking cognitive agents,'' in \emph{CHI'14
  Extended Abstracts on Human Factors in Computing Systems}, 2014, pp.
  2167--2172.

\bibitem{savery_finding_2019}
R.~Savery, R.~Rose, and G.~Weinberg, ``\BIBforeignlanguage{en}{Finding
  {Shimi}'s voice: fostering human-robot communication with music and a
  {NVIDIA} {Jetson} {TX}2},'' \emph{\BIBforeignlanguage{en}{Proceedings of the
  17th Linux Audio Conference}}, p.~5, 2019.

\bibitem{savery2019establishing}
------, ``Establishing human-robot trust through music-driven robotic emotion
  prosody and gesture,'' in \emph{2019 28th IEEE International Conference on
  Robot and Human Interactive Communication (RO-MAN)}.\hskip 1em plus 0.5em
  minus 0.4em\relax IEEE, 2019, pp. 1--7.

\bibitem{sacharin2012geneva}
V.~Sacharin, K.~Schlegel, and K.~Scherer, ``Geneva emotion wheel rating study
  (report). geneva, switzerland: University of geneva,'' \emph{Swiss Center for
  Affective Sciences}, 2012.

\bibitem{coyne2020using}
\BIBentryALTinterwordspacing
A.~K. Coyne, A.~Murtagh, and C.~McGinn, ``Using the geneva emotion wheel to
  measure perceived affect in human-robot interaction,'' in \emph{Proceedings
  of the 2020 ACM/IEEE International Conference on Human-Robot Interaction},
  ser. HRI ’20.\hskip 1em plus 0.5em minus 0.4em\relax New York, NY, USA:
  Association for Computing Machinery, 2020, p. 491–498. [Online]. Available:
  \url{https://doi.org/10.1145/3319502.3374834}
\BIBentrySTDinterwordspacing

\bibitem{walbott98}
H.~G. Walbott, ``Bodily expression of emotion,'' \emph{European Journal of
  Social Psychology}, vol.~28, no.~6, pp. 879 -- 896, 1998.

\bibitem{bretan2015emotionally}
M.~Bretan, G.~Hoffman, and G.~Weinberg, ``Emotionally expressive dynamic
  physical behaviors in robots,'' \emph{International Journal of Human-Computer
  Studies}, vol.~78, pp. 1--16, 2015.

\end{thebibliography}

\end{document}